\pdfoutput=1

\documentclass[11pt]{article}

\usepackage[]{acl}

\usepackage{times}
\usepackage{latexsym}

\usepackage[T1]{fontenc}

\usepackage[utf8]{inputenc}

\usepackage{microtype}

\usepackage{microtype}
\usepackage{amssymb}
\usepackage{xspace}
\usepackage{hhline}
\usepackage{tabularx}
\usepackage{multirow}
\usepackage{adjustbox,booktabs}
\usepackage{mathtools}
\usepackage[multiple]{footmisc}
\usepackage{footnote}
\makesavenoteenv{tabular}
\makesavenoteenv{table}

\usepackage{cleveref}
\crefformat{footnote}{#2\footnotemark[#1]#3}

\newcommand{\eqnref}[2][]{Eqn#1.~(\ref{#2})\xspace}

\newcommand{\ex}[1]{\textit{#1}\xspace}

\title{Unsupervised Lexical Substitution with Decontextualised Embeddings}

\author{Takashi Wada$^{1,2}$ 
\, Timothy Baldwin$^{1,3}$
\,  Yuji Matsumoto$^{2}$
\,  Jey Han Lau$^{1}$ \\
    $^1$ School of Computing and Information Systems, The University of Melbourne \\
    $^2$ RIKEN Center for Advanced Intelligence Project (AIP) \\
    $^3$ Department of Natural Language Processing, MBZUAI \\[0ex]
    \texttt{twada@student.unimelb.edu.au}, \enspace \texttt{tb@ldwin.net}\\\texttt{yuji.matsumoto@riken.jp},\enspace \texttt{jeyhan.lau@gmail.com}
    }    
\begin{document}
\maketitle
\begin{abstract}
We propose a new unsupervised method for lexical substitution using pre-trained language models. Compared to previous approaches that use the generative capability of language models to predict substitutes, our method retrieves substitutes based on the similarity of contextualised and \textit{decontextualised} word embeddings, i.e.\ the average contextual representation of a word in multiple contexts. We conduct experiments in English and Italian, and show that our method substantially outperforms strong baselines and establishes a new state-of-the-art without any explicit supervision or fine-tuning. We further show that our method performs particularly well at predicting low-frequency substitutes, and also generates a diverse list of substitute candidates, reducing morphophonetic or morphosyntactic biases induced by article--noun agreement.\footnote{Code is available at: \url{https://github.com/twadada/lexsub_decontextualised}.}
\end{abstract}

\section{Introduction}
\label{intro}

There has been growing interest in developing automatic writing support systems to assist humans to write documents. One relevant task to this research goal is lexical substitution, where given a target word and its surrounding context, a system suggests a list of word substitutions that can replace the target word without changing its core meaning. For instance, given the target word \ex{{great}} and the context \ex{He is a \underline{great} artist}, the model might suggest alternative words such as \ex{outstanding}, \ex{terrific}, or \ex{distinguished}. Writers can use such suggestions to improve the fluency of their writing, reduce lexical repetition, or search for better expressions that represent their ideas more creatively. 

As with other NLP tasks, recent studies have shown that masked language models such as BERT \cite{bert} perform very well on lexical substitution, even without any task-specific supervision. A common approach is to employ language models as generative models and predict substitutes based on their generative capability. However, this approach has some limitations. First, it is extremely difficult for language models to predict rare words --- especially those that are segmented into multiple subword tokens --- since the models inevitably assign them very low probabilities. Second, word prediction is highly affected by morphosyntactic constraints from the surrounding context, which overshadows the (arguably more important) question of semantic fit. For instance, if the target word is \ex{increase} in the context \ex{... with an \underline{increase} in ...}, language models tend to suggest words that also start with a vowel sound due to the presence of \ex{an} before the target word, missing other possible substitutes such as \ex{hike} or \ex{boost}. In fact, this problem is even more pronounced in languages where words have grammatical gender (e.g.\ Italian nouns) or a high degree of inflection (e.g.\ Japanese verbs).

In this paper, we propose a new approach that explicitly deals with these limitations. Instead of generating words based on language model prediction, we propose to find synonymous words based on the similarity of contextualised and \textit{decontextualised} word embeddings, where the latter refers to the ``average'' contextual representation of a word in multiple contexts. Experiments on English and Italian lexical substitution show that our fully unsupervised method outperforms previous models by a large margin. Furthermore, we show that our model performs particularly well at predicting low-frequency words, and also generates more diverse substitutes with less morphophonetic or morphosyntactic bias, e.g.\ as a result of article--noun agreement in English and Italian.

\section{Method}

\subsection{Our Approach}
Given a sentence that contains a target word $x$ and its surrounding context $c$, we first feed the sentence into a pre-trained transformer model \cite{transformer} such as BERT and generate the contextualised representations of $x$: ${f^{\ell}(x, c)} \in \mathbb{R}^{d}$, where $\ell  ~(\leq L)$ denotes the layer of the model. We propose to predict substitutes of $x$ by retrieving words that have similar representations to ${f^{\ell}(x, c)}$. To this end, we calculate $S(y|x,c)$: the score of $y$ being a substitute of $x$ in context $c$, as follows:
\begin{align}
S(y|x,c)&= \mathrm{cos}(f(y),f(x, c)),\label{eqn_ours}\\
f(x,c) &=\sum_{\ell \in Z}{f^{\ell}(x,c)}, \nonumber \\
f(y) &= \dfrac{1}{N}\sum_{i}^N\sum_{\ell \in Z}{f^{\ell}(y, c'_i)}, \nonumber 
\end{align}
where $f(y) \in \mathbb{R}^{d}$ denotes the \textit{decontextualised} word embedding of $y$; $Z$ is a set of selected layers; and $ \mathrm{cos}(a, b)$ denotes the cosine similarity between $a$ and $b$. To obtain $f(y)$, we randomly sample $N$ sentences ($c'_{1},c'_{2} ...,c'_{N}$) that contain $y$ from a monolingual corpus, and take the average of the contextualised representations of $y$ given $c'_{i}$: $f^{\ell}(y, c'_i)$. We pre-compute $f(y)$ for each word $y$ in our pre-defined vocabulary $\tilde{V}$, which consists of \textit{lexical items} (i.e.\ no subwords) and contains different words from the pretrained model's original vocabulary $V$. If $y$ is segmented into multiple subwords (using the pretrained model's tokeniser), we average its subword representations --- this way we can include low-frequency words in $\tilde{V}$ and generate diverse substitutes. We obtain $f(x,c)$ and  $f(y)$ by summing representations across different layers $\ell \in Z$ to capture various lexical information.\footnote{We also tried taking the weighted sum of the different-layer embeddings, but we did not see noticeable improvement.} 

\subsection{Multi-Sense Embeddings}\label{multi-sense}
Representing $f(y)$ in \eqnref{eqn_ours} as a simple average of the contextualised representations of $y$ is clearly limited when $y$ has multiple meanings, since the representations will likely vary depending on its usage. For instance, \citet{wiedemann-etal-2019-does} show that BERT representations of polysemous words such as \ex{bank} create distinguishable clusters based on their usages. To address this issue, we first group the $N$ sentences into $K$ clusters based on the usages of $y$, and for each cluster $k$, we obtain the decontextualised embedding $f^{k}(y)$ by averaging the contextualised representations, i.e.,
\begin{gather*}
f^{k}(y) = \dfrac{1}{|{C}^k|}\sum_{  c' \in {C}^k}\sum_{ \ell \in Z}{f^{\ell}(y,{c'})},
\label{kmeans}
\end{gather*}
where $C^k$ denotes the set of the sentences that belong to the cluster $k$. To obtain clusters, we apply $K$-means \cite{kmeans,kmeans_pp} to the L2-normalised representations of $y$ in $N$ sentences.\footnote{We concatenate $f^{\ell}(y,{c})$ across multiple layers $\ell \in Z$.} We expect that if $y$ has multiple senses, $f^k{(y)}$ will to some degree capture the different meanings.\footnote{Note that the number of clusters $K$ is fixed across all words, forcing the model to ``split'' and ``lump'' senses \cite{lexicography} to varying degrees.} This methodology has been shown to be effective by \citet{chronis-erk-2020-bishop} on context-independent word similarity tasks. With $f^{k}(y)$, we can refine the similarity score $S(y|x,c)$ in \eqnref{eqn_ours} as follows:
\begin{gather*}
{S}(y|x,c)=\underset{k}{\mathrm{max~}}{\mathrm{cos}(f^{k}(y),f(x, c))}.\label{eqn_ours_k}
\end{gather*}
In this way, we can compare $x$ with $y$ based on the sense that is most relevant to $x$. Furthermore, we capture \ex{global similarity} between $x$ and $y$ as:
\begin{gather}
\begin{aligned}
S(y|x,c) = & \underset{k}{\mathrm{~max~}}\lambda \mathrm{cos}({f^{k}(y),f(x, c)})\\&+(1-\lambda)\mathrm{cos}(f^{k}(y),f^{j_c}(x)),\label{eqn_ours_all}\\
\end{aligned}\\
\begin{aligned}
j_c = \underset{j}{\mathrm{~argmax~}} \mathrm{cos}(f^{j}(x),f(x, c)),\label{k_retrieve}
\end{aligned}
\end{gather}
where the second term in \eqnref{eqn_ours_all} corresponds to the global similarity, which compares $x$ and $y$ outside of context $c$.\footnote{To obtain $f^{j}(x)$, we compute the decontextualised embedding of $x$ and apply $K$-means, as we do to compute $f^{k}(y)$. When $x$ is not included in our pre-defined vocabulary $\tilde{V}$, we set $\lambda$ to 1 and ignore the second term in \eqnref{eqn_ours_all}.} However, it still considers $c$ in \eqnref{k_retrieve} to retrieve the cluster that best represents the meaning of $x$ given $c$.

While \eqnref{eqn_ours_all} generally generates high-quality substitutes, we  found that it sometimes retrieves words that share the same root word as $x$ and yet do not make good substitutes (e.g.\ \ex{pay} and \ex{payer}). This is mainly due to the fact that the vocabulary $\tilde{V}$ contains a large number of derivationally-related words, some of which are out-of-vocabulary (OOV) in the original vocabulary $V$ (e.g.\ \ex{pay~\#\#er}). To address this problem, we add a simple heuristic where $y$ is discarded if the normalised edit distance\footnote{The distance normalised by the maximum string length.} between $x$ and $y$ is less than a threshold (0.5 for our English and Italian experiments).\footnote{We tuned this threshold based on English development data (i.e.\ the development split of SWORDS).}

\subsection{Reranking}
In \eqnref{eqn_ours_all}, the context $c$ affects the representation of $x$ but not $y$. Ideally, however, we want to consider the context $c$ on both sides to find the words that best fit the context. Therefore, we first generate top-$M$ candidates based on \eqnref{eqn_ours_all}, and rerank them using the following score:
\begin{gather} 
\mathrm{S(y|x, c)} = \frac{1}{|Z|}\sum_{\ell \in Z}{\mathrm{cos}(f^{\ell}(y, c),f^{\ell}(x, c))},\label{outs_rerank}
\end{gather}
where $f^{\ell}(y, c)$ denotes the contextualised representation of $y$ given $c$, which can be obtained by replacing $x$ in $c$ with $y$ and feeding it into the model. In \eqnref{outs_rerank}, we calculate the similarity at each layer $\ell\in Z$ and take the average, which yields small yet consistent improvements over averaging the embeddings first and then calculating the similarity.\footnote{In \eqnref{eqn_ours}, we obtained similar results by averaging the embeddings or cosine similarities across layers.} We limit the use of this scoring method to the $M$ candidates only, since it is computationally expensive to calculate $f^{\ell}(y, c)$ for every single word $y$ in $\tilde{V}$. Previously, a similar method was employed by \citet{swords} but they used the last layer only (i.e.\ $Z = \{L\}$). We show that using multiple layers substantially improves the performance. Following \citet{swords}, we set $M$ to 50.

\subsection{Comparison to Previous Approaches}\label{previous_approach}
Our approach contrasts with previous approaches \cite{zhou-etal-2019-bert,swords,tracing} that employ BERT as a generative model and predict lexical substitutes based on the generation probability $ P(y|x,c)$:
\begin{gather}
P(y|x,c) =  \dfrac{\exp(E_y{f^{\hat{L}}(x, c)}+b_y)}{\sum_{\Acute{y} \in V}\exp(E_{\Acute{y}}{f^{\hat{L}}(x, c)}+b_{\Acute{y}})}, \label{eqn_baseline}
\end{gather}
where $E_{y} \in \mathbb{R}^{d}$ denotes the output embedding of $y$, which is usually tied with the input word embedding; ${f^{\hat{L}}(x, c)}$ is the representation at the very last layer of the model;\footnote{Note that this does not always correspond to the last layer of transformer: ${f^{L}(x, c)}$. E.g., BERT calculates ${f^{\hat{L}}(x, c)}$ by applying a feed forward network and layer normalisation to ${f^{L}(x, c)}$, whereas for XLNET, ${f^{\hat{L}}(x, c)}$ = ${f^{L}(x, c)}$.}\footnote{When $x$ consists of multiple subwords, the representation of the first or longest token is usually used.} and $b_y$ is a scalar bias. While this approach is straightforward and well motivated, its predictions are highly influenced by morphosyntax, as discussed in \Cref{intro}. Moreover, \eqnref{eqn_baseline} shows three additional limitations compared to our approach: (1) the prediction is conditioned on the last layer only, despite previous studies showing that different layers capture different information, with the last layer usually containing less semantic information than the lower or middle layers \cite{bommasani-etal-2020-interpreting,tenney-etal-2019-bert}; (2) $y$ is represented by the single vector $E_y$, which may not work well when $y$ has multiple meanings --- we alleviate this by clustering the embeddings (\Cref{multi-sense}); and (3) the model is not capable of generating OOV words, unless we force the model to decode multiple subwords, e.g.\ by using multiple mask tokens or duplicating $x$. Our approach, in contrast, can include rare words in the pre-defined vocabulary $\tilde{V}$ and generate diverse substitutes (\Cref{vocab_effect}).

\section{Experiments}
\subsection{Data and Evaluation}

We conduct experiments in two evaluation settings: generation and ranking. In the generation setting, systems produce lexical substitutes given target words and sentences, while in the ranking setting, they are also given substitute candidates and rank them based on their appropriateness.

For the generation task, we base our experiments on SWORDS \cite{swords}, the largest English lexical substitution dataset, which extends and improves CoInCo \cite{coinco} by introducing a new annotation scheme: in CoInCo, the annotators were asked to come up with substitutes by themselves, whereas in SWORDS, the annotators were given substitute candidates pre-retrieved from a thesaurus, and only had to made binary judgements (``good'' or ``bad'').\footnote{The annotators were asked if they would consider using the substitute candidate to replace the target word as the author of the context.} A word is regarded as {\it acceptable} if it is judged to be good by more than five out of ten annotators, and {\it conceivable} if selected by at least one annotator. In this way, SWORDS provides more comprehensive lists of substitutes, including many low-frequency words that are good substitutes and yet difficult for humans to suggest --- these words are of particular interest to us. For the evaluation metrics, the authors use the harmonic mean of the precision and recall given the gold and top-10 system-generated substitutes.\footnote{More precisely, their evaluation script lemmatises the top-50 substitutes first and then extracts the top-10 distinct words.} As gold substitutes, they use either the acceptable or conceivable words, and calculate the corresponding scores $F_a$ and $F_c$, respectively. They also propose to measure those scores in both {\it \textbf{strict}} and {\it \textbf{lenient}} settings, which differ in that in the lenient setting, candidate words that are not scored under SWORDS are filtered out and discarded.

In the ranking task, we evaluate models on the traditional SemEval-2007 Task 10 (``SemEval-07'') data set (trial+test) \cite{semeval}, as well as SWORDS. For the evaluation metric, we follow previous work in using Generalized Average Precision (GAP; \citet{gap}): 
\begin{gather}
GAP =  \dfrac{\sum_{i=1}^{N}I(\alpha_i)p_i}{\sum_{i=1}^{R}I(\beta_i)\bar{\beta}_i},~~~ p_i = \dfrac{\sum_{k=1}^{i}\alpha_k}{i}, \label{gap_eqn}
\end{gather}
where $\alpha_i$ and $\beta_i$ denote the gold weight of the $i$-th item in the predicted and gold ranked lists respectively, with $N$ and $R$ indicating their sizes; $I(\alpha_i)$ is a binary function that returns $1$ if $\alpha_i>0$, and $0$ otherwise; and $\bar{\beta}_i$ is the average weight of the gold ranked list from the 1st to the $i$-th items. In our task, the weight corresponds to the aptness of the substitute, which is set to zero if it is not in the gold substitutes. Following previous work \cite{melamud-etal-2015-simple,arefyev-etal-2020-always}, we ignore multiword expressions in SemEval-07.\footnote{We run the evaluation code at \url{https://github.com/orenmel/lexsub} with the \ex{no-mwe} option.}

\subsection{Models}

As shown in \eqnref{eqn_ours_all}, our approach requires only the vector representations of words and hence is applicable to any text encoder model. Therefore, we test our method with various pre-trained models, including five masked language models: BERT \cite{bert}, mBERT, SpanBERT \cite{spanbert}, XLNET \cite{xlnet}, and MPNet \cite{mpnet}; one encoder-decoder model: BART \cite{bart}; and two discriminative models: ELECTRA \cite{electra} and DeBERTa-V3 \cite{deberta-v3}.\footnote{See \Cref{model_source} for the details of all models.} We also evaluate two sentence-embedding models: MPNet-based sentence transformer \cite{sentence-bert} and SimCSE \cite{simcse}, both of which are fine-tuned on semantic downstream tasks such as MNLI and achieve good performance on sentence-level tasks. Finally, we also evaluate the encoder of the fine-tuned mBART on English-to-Many translation \cite{tang-etal-2021-multilingual}. Note that the discriminative models and embedding models (e.g.\ NMT-encoder, SimCSE) cannot generate words and hence are incompatible with the previous approach described in \Cref{previous_approach}.
\begin{table}[t]
\begin{center}
\begin{adjustbox}{max width=\columnwidth}
\begin{tabular}{ccc@{\;}ccc@{\;}cc@{\;}}
\toprule
\multirow{2}{*}{Models} &\multicolumn{2}{c}{Lenient}&&\multicolumn{2}{c}{{Strict}}\\
  \cmidrule{2-3}
  \cmidrule{5-6}
  &$F_a$&$F_c$ && $F_a$&$F_c$\\\midrule
\multirow{1}{*}{HUMANS} &\textit{48.8}&\textit{77.9}&&--&-- \\
\multirow{1}{*}{CoInCo} &\textit{34.1}&\textit{63.6}&&--&-- \\
\midrule
\multirow{1}{*}{GPT-3} &\textbf{34.6}& 49.0 &&22.7 &{36.3} \\
\multirow{1}{*}{BERT-K\cref{note1}} &32.4&55.4&&19.2&30.4 \\
\multirow{1}{*}{(w/o rerank)\footnote{\label{note1}These results differ slightly from the reported scores in \citet{swords}, due to a bug in their code.}} &31.8&54.9&&15.7&24.4\\

\multirow{1}{*}{BERT-M} &30.9&48.3&& 16.2& 25.4 \\
\multirow{1}{*}{(w/o rerank)} &30.9 &48.1&& 10.7 &16.5 \\

\multirow{1}{*}{\citet{zhou-etal-2019-bert}\cref{note1}}&32.0&55.4&&17.4&27.5 \\

\multirow{1}{*}{\citet{tracing}\footnote{Updated from the original scores by the authors after they fixed some critical issues in their evaluation setup.}} &{31.9}& 54.9&&16.7 &28.4 \\
\midrule
\multicolumn{6}{c}{OURS}\\\midrule
\multirow{1}{*}{BERT}&33.2&64.1&&21.1&34.9\\

\multirow{1}{*}{(w/o rerank)} &33.0&63.8&&20.7&34.4 \\

\multirow{1}{*}{(w/o rerank, heuristic)} &33.6&64.0&&20.2&32.4 \\\midrule

\multirow{1}{*}{mBERT}&27.0&52.7&&12.4&22.6\\
\multirow{1}{*}{SpanBERT}&32.6&61.4&&20.9&34.0\\
\multirow{1}{*}{MPNet}&33.8&63.8&&22.0&34.1\\
\multirow{1}{*}{XLNet} &34.4&65.3&&23.3&37.4\\
\multirow{1}{*}{ELECTRA} &33.5&64.2&&23.2&36.7\\
\multirow{1}{*}{DeBERTa-V3} &33.6&\textbf{65.8}&&\textbf{24.5}&\textbf{39.9}\\

\multirow{1}{*}{BART (Enc)} &33.6&62.8&&21.9&34.8\\
\multirow{1}{*}{BART (Dec)}&33.5&60.5&&21.4&34.0\\
\multirow{1}{*}{BART (Enc-Dec)} &33.7&64.9&&23.5&37.2\\
\multirow{1}{*}{SBERT (MPNet)} &\textbf{34.6}&64.0&&21.8&33.5\\
\multirow{1}{*}{SimCSE (BERT)} &33.4&64.3&&21.6&35.7\\
\multirow{1}{*}{NMT (mBART)} &28.7&55.6&&13.4&22.2\\\midrule
\multicolumn{6}{c}{OURS (Rank Candidates)}\\\midrule
\multirow{1}{*}{XLNet} &35.2&72.9&&--&-- \\
\multirow{1}{*}{BART} &34.8&72.4&&--&-- \\
\multirow{1}{*}{DeBERTa-V3} &35.1&72.2&&--&-- \\
\bottomrule
\end{tabular}
\end{adjustbox}
\end{center}
\caption{The results for the generation task in the lenient and strict settings. The best scores are boldfaced.}
\label{result_swords}
\end{table}

We use the same vocabulary $\tilde{V}$ for all models, which consists of the 30,000 most common words\footnote{We discard tokens that contain numerals, punctuation, or capital letters. As such, $\tilde{V}$ includes more lexical items (with less noise and no subwords) than the original vocabulary $V$.} in the OSCAR corpus \cite{OSCAR}. We set the number of sentences we sample from OSCAR to calculate the decontextualised embeddings, i.e.\ $N$, to 300; the clustering size $K$ to 4; and $\lambda$ in \eqnref{eqn_ours_all} to 0.7. For the set of transformer layers, $Z$, we employ all layers except for the first and last two, i.e.~\ \{3,~4,~...,~$L-2$\}. We tune all hyper-parameters on the development split of SWORDS.

\subsection{Results}

\Cref{result_swords} shows the results on SWORDS, along with baseline scores from previous work (with some bug fixes, as noted). The first row, \textbf{HUMANS}, indicates the agreement of two independent sets of annotators on (a subset of) SWORDS, and approximates the upper bound for this task. The second row, \textbf{CoInCo}, shows the accuracy of the gold standard substitutes in CoInCo, which are suggested by human annotators without access to substitution candidates\footnote{Since all of these words are in the substitute candidates of SWORDS, it cannot be evaluated under the strict setting.} --- this approximates how well humans perform when asked to elicit candidates themselves. The remainder of the rows above OURS denote baseline systems, all of which employ generative approaches. The first baseline uses \textbf{GPT-3} \cite{gpt-3}, and achieves the state-of-the-art in the strict setting. It generates substitutes based on ``in-context learning'', where the model first reads several triplets of target sentences, queries, and gold-standard substitutes retrieved from the development set, and then performs on-the-fly inference on the test set. As such, it is not exactly comparable to the other fully unsupervised models. \textbf{BERT-K} generates substitutes based on \eqnref{eqn_baseline} by feeding the target sentence into BERT, and \textbf{BERT-M} works the same except that the target word is replaced by [MASK]. Both models further rerank the candidates based on \eqnref{outs_rerank}, using the last layer only; we show the performance without reranking as ``w/o rerank'' in \Cref{result_swords}. \citet{tracing} and \citet{zhou-etal-2019-bert} also use BERT to generate substitutes, and rerank them using their own method.

The rows below OURS indicate the performance of our approach using various off-the-shelf models. Our method with BERT substantially outperforms all the BERT-based baselines, even without the edit-distance heuristic (\Cref{multi-sense}) or reranking method (\eqnref{outs_rerank}). The best performing models are DeBERTa-V3, XLNet, and BART (Enc-Dec), all of which outperform the weakly-supervised GPT-3 model by a large margin in the strict setting; and XLNet even outperforms CoInCo in the lenient setting. The last three rows show the performance of the top-3 models when they are given the candidate words and rank them based on \eqnref{outs_rerank}, which emulates how the SWORDS annotators judged the words. The result shows that all the models still lag behind HUMANS, suggesting there is still substantial room for improvement. Interestingly, BART performs best when we average the scores obtained by its encoder and decoder, suggesting each layer captures complementary information.
It is also intriguing to see that the discriminative models (DeBERTa-V3 and ELECTRA) perform much better than BERT, albeit they are not trained to generate words and not compatible with the previous generative approach. The sentence embedding models (SBERT, SimCSE) perform no better than the original models, which contrasts with their strong performance in sentence-level tasks. The multilingual models (mBERT, NMT) perform very poorly, even though the NMT model was fine-tuned on large English-X parallel corpora. 

\begin{table}[t!]
\begin{center}
\begin{adjustbox}{max width=\columnwidth}
\begin{tabular}{cccc@{\;}ccc@{\;}ccc@{\;}}
\toprule
\multirow{1}{*}{Models} &\multicolumn{1}{c}{S-07}&\multicolumn{1}{c}{SW} \\\midrule
\multirow{1}{*}{HUMANS} &
---&\textit{66.2}\\\midrule
\multirow{1}{*}{\citet{arefyev-etal-2020-always} (XLNet)} &
61.3\footnote{The original score reported by \citet{arefyev-etal-2020-always} is 59.6, but we found we could improve this result by appending unscored OOV words to the ranked list in random order.}&---\\
\multirow{1}{*}{\citet{michalopoulos-etal-2022-lexsubcon} (LMs+WN)} &
 60.3&---\\
\multirow{1}{*}{\citet{alasca} (BERT)}&
58.2&---\\
\multirow{1}{*}{\citet{alasca} (BERT, sup)} &
60.5&---\\
\multirow{1}{*}{\citet{zhou-etal-2019-bert}} (BERT) &
60.5\footnote{Similar to \citet{alasca} and \citet{arefyev-etal-2020-always}, we were unable to reproduce this score.}&53.5\cref{note1}\\
\multirow{1}{*}{\citet{swords} (BERT)} &
56.6&56.9\\

\midrule
\multicolumn{1}{c}{OURS (\eqnref{outs_rerank})}\\
\midrule
\multirow{1}{*}{BERT} &
58.6&60.7\\
\multirow{1}{*}{mBERT} &
45.4&52.0\\
\multirow{1}{*}{SpanBERT} &
59.3&60.8\\
\multirow{1}{*}{MPNet} &
61.5&59.5\\
\multirow{1}{*}{XLNet} &{63.8}&\textbf{62.9}\\
\multirow{1}{*}{ELECTRA} &
{64.4}&62.3\\
\multirow{1}{*}{DeBERTa-V3} &
\textbf{65.0}&\textbf{62.9}\\
\multirow{1}{*}{BART (Enc)} &
62.9&61.9\\
\multirow{1}{*}{BART (Dec)} &
62.6&60.8\\
\multirow{1}{*}{BART (Enc-Dec)} &
64.1&62.7\\
\multirow{1}{*}{SBERT (MPNet)} &
61.0&62.5\\
\multirow{1}{*}{SimCSE (BERT)} &
58.4&60.9\\
\multirow{1}{*}{NMT (mBART, Enc)} &
46.0&51.5\\

\bottomrule
\end{tabular}
\end{adjustbox}
\end{center}
\caption{GAP scores on SemEval-07 and SWORDS. ``LMs+WN'' employs multiple language models and WordNet, and ``BERT, sup'' is a supervised model.}
\label{result_gap}
\end{table}

\Cref{result_gap} shows the results for the ranking task on SemEval-07 and SWORDS. \citet{michalopoulos-etal-2022-lexsubcon} harness WordNet \cite{wordnet} to obtain synsets of the target word and also their glosses, and employ BERT and RoBERTa \cite{roberta} to rank candidates. The models proposed by \citet{alasca} are different from the others in that they fine-tune BERT on lexical substitution data sets. They propose unsupervised (BERT) and supervised (BERT, sup) models, which are fine-tuned on automatically-generated or manually-annotated data. \Cref{result_gap} shows that our method with BERT performs comparably with the unsupervised model of \citet{alasca} without any fine-tuning, and outperforms \citet{zhou-etal-2019-bert} and \citet{swords} (except for the score of \citet{zhou-etal-2019-bert} on SemEval-07, which couldn't be reproduced in previous work). Just like the generation task, DeBERTa-V3 achieves the best performance on both data sets and establishes a new state-of-the-art. Other models also follow a similar trend to the generation results, e.g.\ BART performs best by combining its encoder and decoder, and the multilingual models perform very poorly. We hypothesise that their poor performance is mainly caused by suboptimal segmentation of English words. This hypothesis is also supported by the fact that DeBERTa-V3 has by far the largest vocabulary $V$ of all models.\footnote{Note that $V$ differs from $\tilde{V}$, the pre-defined vocabulary we used for all models. \Cref{model_source} compares the size of the model's original vocabulary $V$ across different models.}
 
\subsection{Results on Italian Lexical Substitution}

We further conduct an additional experiment on Italian, based on the data set from the EVALITA 2009 workshop \cite{evalia_2009}. We report $F$ scores given top-10 predictions as in the English generation task, plus two traditional metrics used in the workshop, namely {\it oot} and {\it best}, which compare the top-10 and top-1 predictions against the gold substitutes.\footnote{We report precision only, as it is the same as recall under those metrics when predictions are made for every sentence.} We lemmatise all the generated words to make them match the gold substitutes, following the SWORDS evaluation script.\footnote{We used the Italian lemmatiser (it\_core\_news\_sm 3.2.0) in spaCy (ver. 3.2.2) \cite{spacy}.} We use the same hyper-parameters as for the English experiments, and \Cref{Italian} shows the results. \citet{hintz-biemann-2016-language} is a strong baseline that retrieves substitute candidates from MultiWordNet \cite{multiwordnet} and ranks them using a supervised ranker model. We also implement BERT-K using an Italian BERT model \cite{italian_bert}, with and without the reranking method. The results show that our approach substantially outperforms the baselines, confirming its effectiveness. However, our reranking method is not as effective as in English, which we attribute to the influence of grammatical gender in Italian (which we return to in \Cref{analysis_syntax}). The heuristic improves best-P but harms $F$ and oot-P, meaning it removes good candidates as well as bad ones, possibly because we used the threshold tuned on English.

\begin{table}[t!]
\begin{center}
\begin{adjustbox}{max width=\columnwidth}
\begin{tabular}{cccccc}
\toprule
\multirow{1}{*}{} &$F$&\multicolumn{1}{c}{best-P}&\multicolumn{1}{c}{oot-P}\\\midrule
 \multirow{1}{*}{\citet{hintz-biemann-2016-language}}  &---& 16.2&41.3\\
  \multirow{1}{*}{BERT-K}  &14.3&14.4&39.1\\
    \multirow{1}{*}{(w/o rerank)}  &15.6&17.4&43.3\\
 \midrule

\multirow{1}{*}{OURS (BERT)} &17.3&{19.9}&47.5\\
\multirow{1}{*}{(w/o rerank)} &{17.5}&19.1&48.4\\

\multirow{1}{*}{(w/o rerank, heuristic)} &{17.5}&17.4&{48.7}\\\midrule
 \multirow{1}{*}{OURS (ELECTRA)} &{19.0}&{21.0}&{51.2}\\
  \multirow{1}{*}{(w/o rerank)} &{18.9}&\textbf{21.3}&{51.0}\\
\multirow{1}{*}{(w/o rerank, heuristic)} &\textbf{19.2}&{20.2}&\textbf{52.1}\\

\bottomrule
\end{tabular}
\end{adjustbox}
\end{center}
\caption{The result of Italian lexical substitution.}
\label{Italian}
\end{table}

\begin{table}[t!]
\begin{center}
\begin{adjustbox}{max width=\columnwidth}
\begin{tabular}{cccccc}
\toprule
\multirow{1}{*}{} &BERT&BART&XLNet&DeBERTa\\\midrule
\multirow{1}{*}{$\lambda=1$} &34.1&26.2&32.9&35.8 \\
\multirow{1}{*}{$\lambda=0$} &32.8&34.1&34.1&35.6 \\
  \multirow{1}{*}{$K = 1$} &32.9&32.0&33.7&35.7 \\
\multirow{1}{*}{$k$ is random} &30.6&29.2&30.3&32.0 \\
\multirow{1}{*}{w/o heuristic} &32.4&32.0&33.4&35.7 \\

\midrule
\multirow{1}{*}{$\lambda=0.7$} &34.4&34.0&35.0&36.9 \\
\multirow{1}{*}{+ rerank} &\textbf{34.9}&\textbf{37.2}&\textbf{37.4}&\textbf{39.9} \\

\bottomrule
\end{tabular}
\end{adjustbox}
\end{center}
\caption{Ablation studies of our method. The scores denote $F_c$ in the strict setting on SWORDS.}
\label{ablation}
\end{table}

\section{Analysis}
\subsection{Ablation Studies}\label{ablation_sec}
We perform ablation studies on SWORDS to see the effect of $\lambda$ and the $K$-clustered embeddings, and also the heuristic based on edit distance. \Cref{ablation} shows the results. Overall, our method with $\lambda = 0.7$ performs better than $\lambda = 1$ or $\lambda = 0$, confirming the benefit of considering both in-context and out-of-context similarities. One interesting observation is that while BERT and DeBERTa perform better with $\lambda = 1$ than with $\lambda = 0$, the opposite trend is observed for XLNet and especially BART (and hence the optimal value for $\lambda$ is smaller than 0.7). This suggests that BART representations are highly influenced by context, containing much information that is not relevant to the semantics of the target word; we further confirm this in the next section. When we set the cluster size $K$ to 1, the performance of all the models drops sharply, indicating the effectiveness of the clustered embeddings. When we retrieve the cluster of $y$ at random instead of the closest one to $x$ in \eqnref{eqn_ours_all}, the performance decreases substantially, suggesting each cluster captures different semantics. The heuristic consistently improves the performance, filtering out derivationally-related yet semantically-dissimilar words to the target word. Lastly, our reranking method substantially improves the performance of all the models, demonstrating that it is important to incorporate the target context $c$ into both the target and candidate word representations.

\subsection{Effects of Morphosyntactic Agreement}\label{analysis_syntax}

\begin{table}[t]
\begin{center}
\begin{adjustbox}{max width=\columnwidth}
\begin{tabular}{ccccccc}
\toprule
 &\ex{a}&\ex{an}&\ex{un}&\ex{una}&\ex{la/le}&\ex{il/i}\\\midrule

    \multirow{1}{*}{BERT-K}  &94.2&56.0&93.3&92.5&93.8&92.0\\
    \multirow{1}{*}{(w/o r)}  &91.1&54.0&90.0&87.5&90.3&87.4\\\midrule

    \multirow{1}{*}{OURS (BERT)}  &88.4&\underline{24.0}&81.9&89.2&92.6&85.2\\        \multirow{1}{*}{(w/o r)}  &\underline{86.8}&18.0&70.5&66.7&69.7&\underline{64.6}\\
        \multirow{1}{*}{(w/o r, h)} &\underline{86.8}&44.0&\underline{62.4}&\underline{65.0}&\underline{69.4}&60.7\\\midrule

  \multirow{1}{*}{Gold}  &86.8&26.9&55.6&63.3&69.0&67.0\\

\bottomrule
\end{tabular}
\end{adjustbox}
\end{center}
\caption{The percentage of generated and gold substitutes whose initial sound or grammatical gender agrees with the corresponding English or Italian articles. ``r'' and ``h'' denote rerank and heuristic, respectively. The closest numbers to Gold are underlined.}
\label{result_article}
\end{table}

Compared to previous generative approaches, our method does not depend on the generation probabilities of language models, and hence we expect it to be less sensitive to morphosyntactic agreement effects. To investigate this, we analyse the performance on noun target words which immediately follow one of the following articles: \ex{a} or \ex{an} in English, and \ex{una}, \ex{la}, \ex{le}, \ex{un}, \ex{il} or \ex{i} in Italian. The first three Italian articles are used with feminine nouns, and the rest with masculine ones. Our hypothesis is that generative methods will be highly biased by these articles, despite the gold standard being \textit{semantically} annotated, and thus largely oblivious to local morphosyntactic agreement effects.

\Cref{result_article} shows the percentage of top-10 predicted candidates that agree with the article.\footnote{We retrieve Italian gender information using a dictionary API (\url{https://github.com/sphoneix22/italian_dictionary}), and English phonetic information using CMUdict (\url{https://github.com/cmusphinx/cmudict}) accessed via NLTK \cite{nltk}.} It demonstrates that the prediction of BERT-K is highly affected by the proceeding article as expected, resulting in substitutes which don't satisfy this constraint being assigned low probabilities. In contrast, the results of our method are more balanced and close to the gold standard.\footnote{Note that the big jump in results for \ex{an} is based on a small number of instances (5 sentences).} Conversely, our reranking method actually \textit{increases} the bias greatly, suggesting that the contextualised embeddings $f^{\ell}(x,c)$ and $f^{\ell}(y,c)$ in \eqnref{outs_rerank} become similar when $x$ and $y$ collocate similarly with the words in the context $c$ --- overall, this leads to better results, but actually hurts in cases of local agreement effects biasing the results. This is one reason why reranking was not as effective in Italian as in English, as agreement effects are stronger in Italian. 

\begin{table}[t!]
\begin{center}
\begin{adjustbox}{max width=\columnwidth}
\begin{tabular}{cccc@{\;}ccc@{\;}ccc@{\;}}
\toprule
\multirow{1}{*}{Models} &\multicolumn{1}{c}{\ex{a}}&\multicolumn{1}{c}{\ex{an}} \\\midrule
BERT-K &94.2&56.0\\
BERT-M &99.5&88.0\\
\midrule

BERT &88.4&24.0\\
SpanBERT &97.9&46.0\\
MPNet &91.6&52.0\\
XLNet  &83.2&38.0\\
BART (Enc) &
89.5&62.0\\
BART (Dec) &
91.1&46.0\\
BART (Enc-Dec) &
89.5&58.0\\
DeBERTa-V3&
86.8&32.0\\
ELECTRA&
88.4&36.0\\

\bottomrule
\end{tabular}
\end{adjustbox}
\end{center}
\caption{The percentage of substitutes whose initial sound agrees with the corresponding English articles.}
\label{syntactic_effect_eng}

\end{table}

\begin{table*}[t]
\begin{center}
\begin{adjustbox}{max width=\textwidth}
\begin{tabular}{m{0.22\linewidth}  m{0.78\linewidth}}
\toprule
\multirow{1}{*}{Context}&The loan may be extended by the McAlpine group for an additional year with an \underline{\textbf{increase}} in the conversion price to \$2.50 a share.\\\midrule
\multirow{1}{*}{Gold (Conceivable)} &
\ex{boost}, \ex{gain}, \ex{raise}, \ex{hike}, \ex{rise}, swell, surge, upsurge, enlargement, growth, addition, escalation, expansion, upgrade, cumulation, swelling, exaggeration, step-up
\\\midrule
\multirow{1}{*}{BERT-K} &increased, \textbf{rise}, enhancement, increasing, \textbf{addition} 
\\\midrule

\multirow{1}{*}{OURS (BERT)} &\textbf{rise}, \textbf{raise}, change, reduce, reduction
\\
\multirow{1}{*}{OURS (BART)} &\textbf{rise}, uptick, \textbf{hike}, improvement, upping
\\
\multirow{1}{*}{OURS (DeBERTa-V3)} &\textbf{boost}, \textbf{rise}, \textbf{raise}, \textbf{hike}, reduction
\\\midrule\midrule
\multirow{1}{*}{Context}&Under an \underline{\textbf{accord}} signed yesterday, the government and Union Bank of Finland would become major shareholders in the new company, each injecting 100 million Finnish markkaa (\$23.5 million).\\\midrule
\multirow{1}{*}{Gold (Conceivable)} &
\ex{arrangement}, \ex{agreement}, \ex{pact}, contract, deal, treaty\\\midrule
\multirow{1}{*}{BERT-K} &understanding, \textbf{agreement}, \textbf{pact}, \textbf{arrangement}, agreeing
\\\midrule

\multirow{1}{*}{OURS (BERT)} &\textbf{agreement}, \textbf{pact}, \textbf{treaty}, understanding, \textbf{deal}
\\
\multirow{1}{*}{OURS (BART)} &\textbf{agreement}, \textbf{deal}, \textbf{pact}, \textbf{arrangement}, \textbf{treaty}
\\
\multirow{1}{*}{OURS (DeBERTa-V3)} &
\textbf{pact}, \textbf{agreement}, \textbf{deal}, \textbf{arrangement}, \textbf{treaty}
\\

\bottomrule
\end{tabular}
\end{adjustbox}
\end{center}
\caption{Top-5 predictions when the article \ex{an} comes before the target word (\ex{increase} or \ex{accord}). Gold shows a list of ``conceivable'' words sorted by their annotated scores (with ``acceptable'' words shown in italic, and multiword expressions omitted from the table). Words included in Gold are boldfaced.}
\label{an_examples}
\end{table*}

\Cref{syntactic_effect_eng} shows the result when we use different pre-trained models in English. First, it shows that BERT-M is more sensitive to the articles than BERT-K, indicating the strong morphophonetic agreement effect on the masked word prediction. Among the pre-trained language models used by our method, SpanBERT and BART are the most sensitive to the article \ex{a} and \ex{an}, respectively. This suggests that the embeddings $f(x,c)$ obtained from these models are highly sensitive to the context $c$, partly explaining why BART performs very poorly with $\lambda$ = 1, as shown in \Cref{ablation_sec}. Lastly, \Cref{an_examples} shows examples of predicted substitutes when the article \ex{an} comes before the target word. It shows that BERT-K and OURS with BART tend to retrieve words that start with a vowel sound, as quantitatively described in \Cref{syntactic_effect_eng}.

\begin{figure}[t] 
\begin{center}
\includegraphics[bb=020 30 500 320,scale=0.5]{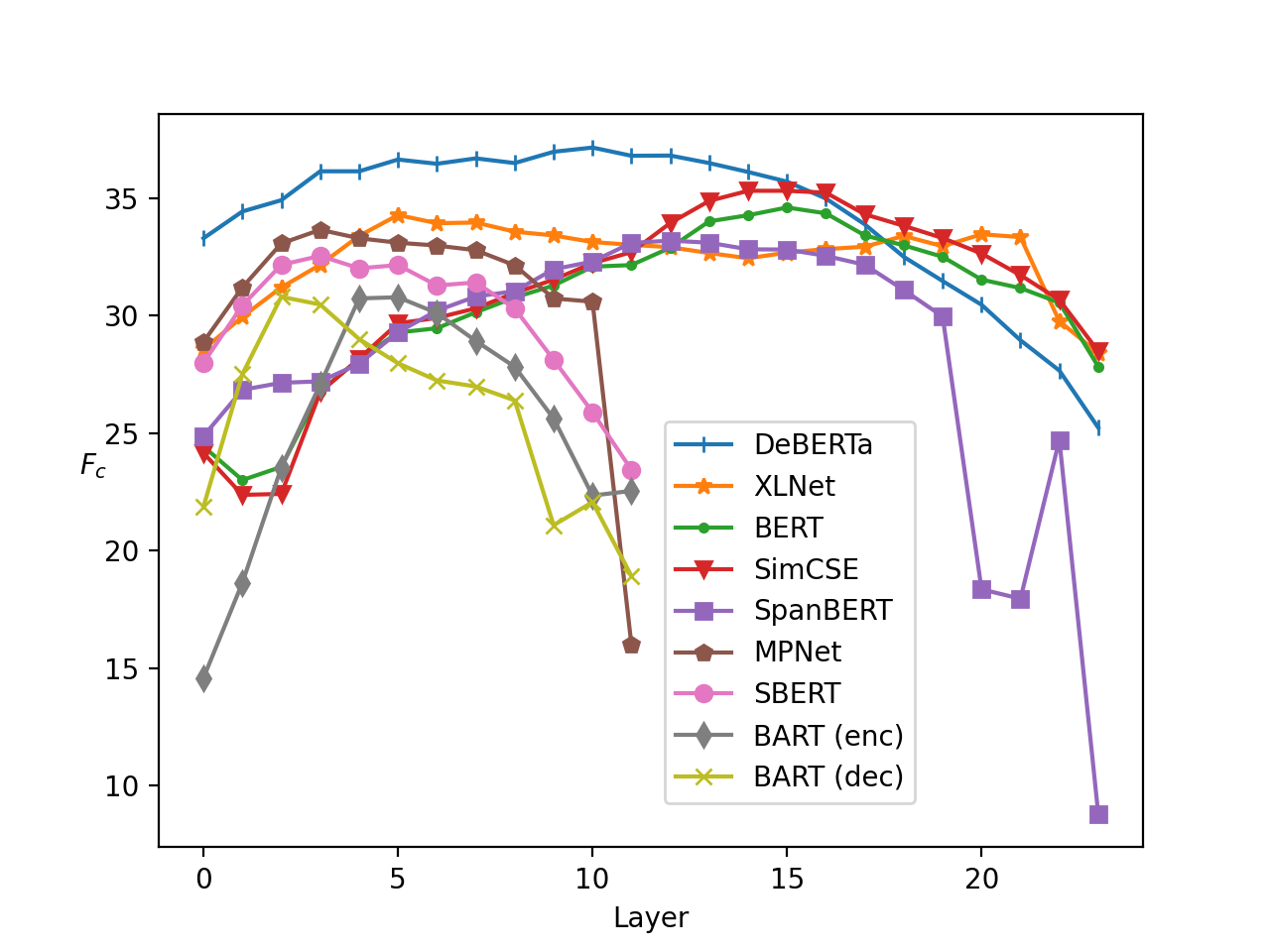}
\end{center}
\caption{Layer-wise performance ($F_c$) on SWORDS.}
\label{layer_fig}
\end{figure}

\subsection{Layer-Wise Performance}\label{layer-analysis}

We analyse the performance on SWORDS using different layers in \Cref{layer_fig} (w/o rerank).\footnote{We perform qualitative analysis in \Cref{layer_examples_section}.} First, we clearly see that middle layers perform better than the first or last ones, for all models.\footnote{We see a similar trend in the ranking task (\Cref{layerwise-ranking-sec}).} The performance of BERT peaks at layer 16, in contrast with previous findings that the first quarter of layers perform best on context-independent word similarity tasks \cite{bommasani-etal-2020-interpreting}, likely because lexical substitution critically relies on context.\footnote{In fact, \citet{tenney-etal-2019-bert} show that high-level semantic information is encoded in higher layers.} Our method using multiple layers performs mostly as well as using the best layer without the need to perform model-wise layer selection (see \Cref{layer-wise-table} in \Cref{multi-layer-sec}). The last layer performs very poorly for all models, highlighting the limitation of the previous approach which uses the last layer only (\eqnref{eqn_baseline}). The downward trend is particularly evident for MPNet, BART (dec), and SpanBERT; for BART and SpanBERT, we attribute this to the fact that their last-layer representations of the word at position $t$ are used to predict the next word $w_{t+1}$, or $u$ neighbouring words \{$w_{t-u}$..,$w_{t-1}$\} or \{$w_{t+1}$..,$w_{t+u}$\}.\footnote{SpanBERT does this for Span Boundary Objective.} This training objective may also lead to their sensitivity to articles before the target word, as shown in \Cref{analysis_syntax}. Interestingly, the sentence-embedding models (SBERT and SimCSE) are no exception to the downward trend, which is somewhat counter-intuitive given that their last layer representations are fine-tuned (and used during inference) to perform semantic downstream tasks. Importantly, they do not perform better than the original models (MPNet and BERT), although in the ranking task, both models benefit moderately from fine-tuning (see \Cref{layerwise-ranking-sec}). 

\begin{table}[t!]
\begin{center}
\begin{adjustbox}{max width=\columnwidth}
\begin{tabular}{cccccc}
\toprule
\multirow{2}{*}{} &\multicolumn{3}{c}{\# Matched Words}&\multirow{2}{*}{$F_{c}$}\\
\cmidrule{2-4}
\multirow{1}{*}{} & low&med&high&&\\\midrule
  \multirow{1}{*}{BERT-K}  & 121& 144 &2002&30.4\\
 \midrule
 \multirow{1}{*}{BERT (5k)}&31&52& 2017&28.1\\
\multirow{1}{*}{BERT (10k)} & 72& 111 &2249&32.6\\
\multirow{1}{*}{BERT (20k)} &164& 220 &2194&34.5\\

\multirow{1}{*}{BERT (30k)} &241&267&2095
&{34.9}\\\midrule
\multirow{1}{*}{BART (30k)}& 380&  274 & 2123& 37.2\\
\multirow{1}{*}{XLNet (30k)}& 395 &\textbf{292} &2106& 37.4\\
\multirow{1}{*}{DeBERTa (30k)}& \textbf{429}&287 &\textbf{2262} &\textbf{39.9}\\

\bottomrule
\end{tabular}
\end{adjustbox}
\end{center}
\caption{The number of correctly predicted substitutes, grouped by their frequency in monolingual data. The number in brackets shows the size of the vocabulary $\tilde{V}$.}
\label{analysis_freq}
\end{table}

\begin{table}[t!]
\begin{center}
\begin{adjustbox}{max width=\columnwidth}
\begin{tabular}{ccccccc}
\toprule
\multirow{1}{*}{}$K$ &1&2& 4&8&16\\\midrule
  \multirow{1}{*}{$F_a$} &22.8&23.1&23.2&23.3&\textbf{23.4}\\
\multirow{1}{*}{$F_{c}$} &36.0& 36.7&36.7&\textbf{36.9}&36.7\\
\bottomrule
\end{tabular}
\end{adjustbox}
\end{center}
\caption{Results with different numbers of clusters.}
\label{cluster_analysis}
\end{table}

\subsection{Analysis of Word Frequency}\label{vocab_effect}

One of the strengths of our approach is that it can generate low-frequency substitutes that are OOV words in the original vocabulary. To confirm this, we analyse how well our method can generate low-frequency words from different vocabulary sizes $\tilde{V}$. \Cref{analysis_freq} shows the results, in which we experiment with our BERT-based model with the vocabulary sizes of 5k, 10k, 20k, and 30k. The columns under ``\# Matched Words'' show the numbers of correctly-predicted words, grouped by frequency range: low, med, and high denote words with frequency $<$50k, 50k--100k, and $>$100k in a large web corpus. The table shows that our method with 30k words generates nearly twice as many low-frequency substitutes as the baseline. Our method with 10k words still outperforms BERT-K in $F_c$, demonstrating its effectiveness. The last three rows show the performance of our method using other models, further demonstrating its ability to predict low-frequency words.

\subsection{Effects of Cluster Size}\label{efffect_cluster}

Finally, we analyse the effect of the cluster size $K$ for ELECTRA, as shown in \Cref{cluster_analysis}. While a larger cluster size yields better performance, the improvement is marginal. Rather than using a fixed $K$, in future work we are interested in dynamically selecting the number of clusters per word.

\section{Related Work}

In the pre-BERT era, most lexical substitution methods employed linguistic resources such as WordNet \cite{wordnet} to obtain substitute candidates \cite{szarvas-etal-2013-supervised,hintz-biemann-2016-language}. However, recent studies have shown that pre-trained language models such as BERT outperform these models without any external lexical resources. For instance, \citet{zhou-etal-2019-bert} feed a target sentence into BERT while partially masking the target word using dropout \cite{dropout}, and generate substitutes based on the probability distribution at the target word position. The masking strategy was shown to be effective on SemEval-07 but not on SWORDS. Similarly, \citet{tracing} feed two sentences into BERT, concatenating the target sentence with itself but with the target word replaced by {[}MASK{]}, and predict words based on the mask-filling probability. \citet{michalopoulos-etal-2022-lexsubcon} augment pre-trained language models with WordNet and outperform \citet{zhou-etal-2019-bert}. \citet{alasca} fine-tune BERT on lexical substitution data sets that are automatically generated using BERT. They show that this approach is effective at ranking, and that adding manually-annotated data further boosts performance. \citet{lacerra-etal-2021-genesis} fine-tune BART on human-annotated data, and make it generate a list of substitutes given a target sentence in an end-to-end manner. They show that this generative approach rivals \citet{zhou-etal-2019-bert}. Note that all of these recent models are evaluated on English only.
\section{Conclusion}
We present a new unsupervised approach to lexical substitution using pre-trained language models. We showed that our method substantially outperforms previous methods on English and Italian data sets, establishing a new state-of-the-art. By comparing performance on lexical substitution using different layers, we found that middle layers perform better than first or last layers. We also compared the substitutes predicted by the previous generative approach and our method, and showed that our approach works better at predicting low-frequency substitutes and reduces morphophonetic or morphosyntactic biases induced by article--noun agreement in English and Italian.

\bibliography{acl_latex}
\bibliographystyle{acl_natbib}
\appendix

\section{Details of Pre-trained Models}
\label{model_source}

\Cref{mdoel_details} describe the details of the pre-trained models used in our experiments. We sourced these models from the Transformers library \cite{wolf-etal-2020-transformers} except for SpanBERT, which we obtained from the original GitHub repository (\url{https://github.com/facebookresearch/SpanBERT}).

\section{Layer-Wise Ranking Performance}\label{layerwise-ranking-sec}

\begin{figure}[t] 
\begin{center}
\includegraphics[bb=020 30 500 320,scale=0.5]{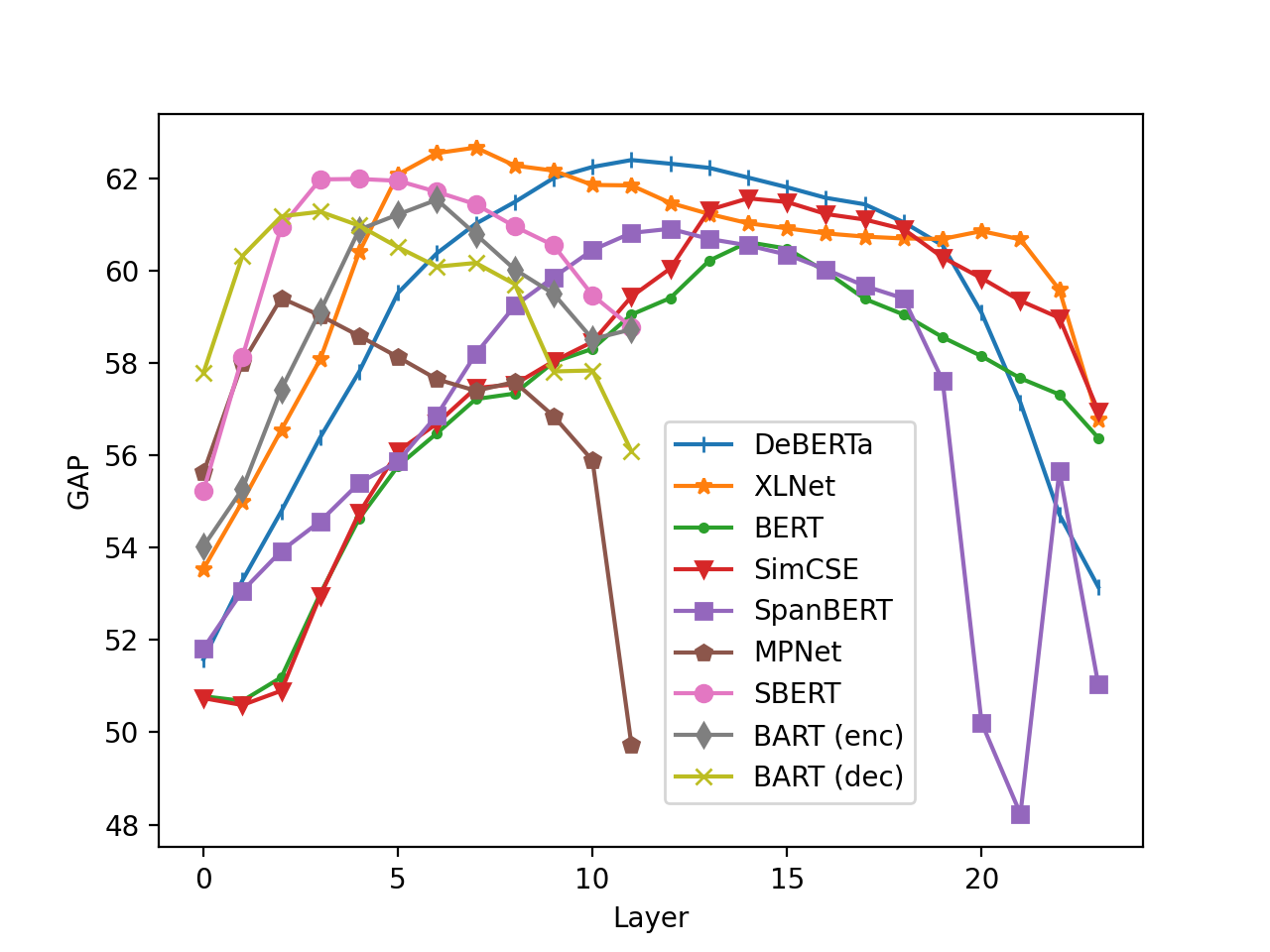}
\end{center}
\caption{Layer-wise performance (GAP) on SWORDS.}
\label{layerwise_rank_fig}
\end{figure}

\Cref{layerwise_rank_fig} shows the layer-wise performance in the ranking task. Similar to the generation results (\Cref{layer_fig}), middle layers perform better than the first or last layers. It also shows that sentence-embedding models (SimCSE/SBERT) outperform their original models (BERT/MPNet) for several layers, different from the generation results where they perform similarly. This suggests that fine-tuning on semantic downstream tasks improves the capacity of the model to differentiate subtle semantic differences between synonymous words, but not their ability to retrieve relevant words from a large pool of words; it also suggests that optimal representations for these objectives might differ.

\section{Effectiveness of Using Multiple Layers}\label{multi-layer-sec}

\Cref{layer-wise-table} shows the generation and ranking performance of our model on SWORDS using different layers. It shows that our method using multiple layers $\ell \in Z$ performs comparably or even better than selecting the best layer tuned on the \textit{test} set for each model. It also shows that the best-performing layer differs across models, suggesting they capture lexical information in a different manner. 

\begin{table*}[t]
\begin{center}
\begin{adjustbox}{max width=\textwidth}
\begin{tabular}{cccc@{\;}ccc@{\;}ccc@{\;}}
\toprule
\multirow{1}{*}{Models} &\multicolumn{1}{c}{\# Layer}&\multicolumn{1}{c}{Emb Size} &$|V|$&\multicolumn{1}{c}{Model Path} \\\midrule
\multirow{1}{*}{BERT} &
24&1024&30522&bert-large-uncased\\
\multirow{1}{*}{mBERT} &
12&768&105879*&bert-base-multilingual-uncased\\
\multirow{1}{*}{SpanBERT} &
24&1024&30522&spanbert-large-cased\\
\multirow{1}{*}{MPNet} &
12&768&30527&microsoft/mpnet-base\\
\multirow{1}{*}{XLNet} &24&1024&32000&xlnet-large-cased\\
\multirow{1}{*}{ELECTRA} &
24&1024&30522&google/electra-large-discriminator\\
\multirow{1}{*}{DeBERTa-V3} &
24&1024&128000&microsoft/deberta-v3-large\\
\multirow{1}{*}{BART (Enc/Dec)} &
12&1024&50265&facebook/bart-large\\
\multirow{1}{*}{SBERT (MPNet)} &
12&768&30527&sentence-transformers/all-mpnet-base-v2\\
\multirow{1}{*}{SimCSE (BERT)} &
24&1024&30522&princeton-nlp/sup-simcse-bert-large-uncased\\
\multirow{1}{*}{NMT (mBART, Enc)} &
12&1024&250054*&facebook/mbart-large-50-one-to-many-mmt\\\midrule

\multirow{1}{*}{BERT (Italian)} &
12&768&31102&dbmdz/bert-base-italian-xxl-uncased\\
\multirow{1}{*}{ELECTRA (Italian)} &
12&768&31102&dbmdz/electra-base-italian-xxl-cased-discriminator\\

\bottomrule
\end{tabular}
\end{adjustbox}
\end{center}
\caption{Details of the pre-trained models used in this paper. $|V|$ denotes the original vocabulary size of each model. *The vocabularies of mBERT and mBART contain a great number of non-English words.}
\label{mdoel_details}
\end{table*}

\begin{table*}[t!]
\begin{center}
\begin{adjustbox}{max width=\textwidth}
\begin{tabular}{ccccccccccccc}
\toprule
&\multicolumn{5}{c}{Generation Performance ($F_c$)}&&\multicolumn{5}{c}{Ranking Performance (GAP)}\\
 
 \cmidrule{2-6}\cmidrule{8-12} 
 
 Layer&\multicolumn{1}{c}{First}&\multicolumn{1}{c}{Middle}&\multicolumn{1}{c}{Last}&\multicolumn{1}{c}{Best}&\multicolumn{1}{c}{$\ell \in Z$}
&&\multicolumn{1}{c}{First}&\multicolumn{1}{c}{Middle}&\multicolumn{1}{c}{Last}&\multicolumn{1}{c}{Best}&\multicolumn{1}{c}{$\ell \in Z$} \\\midrule
\multirow{1}{*}{BERT} &24.4&32.2&27.8&\textbf{34.6} (16) &34.4&&50.8&59.1 &56.4&60.6 (15)&\textbf{60.7}\\
SpanBERT &24.9&33.1&8.8&\textbf{33.2} (13)&31.1&&51.8&60.8&51.0&\textbf{60.9} (13)&60.8\\
MPNet&28.9&33.1&16.0&33.7 (4) &\textbf{33.8}&&55.6&58.1&49.7&59.4 (3)&\textbf{59.5}\\
XLNet  &28.6&33.0&28.4&34.3 (6)&\textbf{35.0}&&53.5&61.9&56.8&62.7 (8)&\textbf{62.9}\\
DeBERTa-V3&33.3&36.8&25.2&\textbf{37.2} (11)&36.9&&51.6&62.4&53.2&62.4 (12)&\textbf{62.9}\\

BART (Enc) &14.6&\textbf{30.8}&22.6&\textbf{30.8} (6)&30.0&&54.0&61.2&58.7&61.5 (7)&\textbf{61.9}\\
BART (Dec) &21.9&28.0&18.9&\textbf{30.8} (3)&29.0&&57.8&60.5&56.1&\textbf{61.3} (4)&60.8\\
\bottomrule
\end{tabular}
\end{adjustbox}
\end{center}
\caption{Generation and ranking performance of our approach on SWORDS using the first, middle ($\frac{L}{2}$th), last ($L$th), or best layer tuned on the \textit{test} set (the corresponding layer denoted in brackets); or using multiple layers $\ell \in Z$:~\{3,~4,~...,~$L-2$\} ($L$~=~12 for MPNet and BART, and 24 otherwise). Generation and ranking performance across all layers is illustrated in \Cref{layer_fig} and \Cref{layerwise_rank_fig}.}
\label{layer-wise-table}

\end{table*}

\begin{table*}[t]
\begin{center}
\begin{adjustbox}{max width=\textwidth}
\begin{tabular}{m{0.18\linewidth}  m{0.82\linewidth}}
\toprule
\multirow{1}{*}{Context}& I say I do not \underline{\textbf{care}} about law, I care about service and she should care about money.\\\midrule
\multirow{1}{*}{Gold (Conceivable)} &\ex{worry}, think, mind, desire, love, tend, cherish, consider, stress, concern, bother, watch\\\midrule
\multirow{1}{*}{BERT (L3-22)} &matter, caregiving, carefree, \textbf{worry}, \textbf{concern}, know, pay, caregivers, \textbf{love}, like\\


\multirow{1}{*}{BART (L3-10)} &\textbf{concern}, matter, carelessness, caretaker, carelessly, carefree, caretakers, \textbf{worry}, \textbf{bother}, \textbf{mind}\\\midrule

\multirow{1}{*}{BERT (L1)} & caregiving, carefree, caregiver, caregivers, aftercare, childcare, healthcare, skincare, custody, affections\\
\multirow{1}{*}{BERT (L12)} &matter, caregiving, \textbf{worry}, carefree, fret, \textbf{love}, despise, loathe, resent, pay\\
\multirow{1}{*}{BERT (L24)} &matter, \textbf{worry}, pay, \textbf{concern}, know, look, give, take, \textbf{bother}, \textbf{think}\\
\midrule



\multirow{1}{*}{BART-Enc (L1)} &aftercare, caregiving, caretaker, carefree, carelessly, carelessness, caretakers, healthcare, caregivers, medicare\\
\multirow{1}{*}{BART-Enc (L6)} &\textbf{concern}, todo, caretaker, interest, carelessness, careless, disinterested, disdain, pay, carelessly\\
\multirow{1}{*}{BART-Enc (L12)} &todo, aswell, beleive, inbetween, zealand, pay, \textbf{concern}, usefull, noone, ofcourse\\
\midrule
\multirow{1}{*}{BART-Dec (L1)} &caregiving, carelessness, caretaker, carelessly, carefree, aftercare, caretakers, \textbf{concern}, healthcare, \textbf{worry}\\
\multirow{1}{*}{BART-Dec (L6)} &\textbf{concern}, \textbf{bother}, \textbf{worry}, commit, reckon, dispose, shit, grieve, pay, strive\\
\multirow{1}{*}{BART-Dec (L12)} &\textbf{worry}, damn, \textbf{concern}, inquire, complain, passionate, shit, \textbf{think}, talk, whine\\
\midrule\midrule
\multirow{1}{*}{Context}&``I’m starting to see more business transactions,'' says Andrea West of American Telephone \& Telegraph Co., noting growing \underline{\textbf{interest}} in use of 900 service for stock sales, software tutorials and even service contracts.\\\midrule
\multirow{1}{*}{Gold (Conceivable)} &\ex{interestedness}, enthusiasm, demand, attraction, popularity, excitement, curiosity, activity, importance, notice, significance, involvement, relevance, note, gain, passion, influence, accrual, concernment
\\\midrule
\multirow{1}{*}{BERT (L3-22)} &\textbf{curiosity}, \textbf{enthusiasm}, intrigued, desire, concern, fascination, \textbf{passion}, attention, \textbf{excitement}, fondness\\
\multirow{1}{*}{BART (L3-10)} &fascination, \textbf{enthusiasm}, appetite, \textbf{curiosity}, \textbf{excitement}, concern, inclination, eagerness, desire, \textbf{involvement}\\\midrule

\multirow{1}{*}{BERT (L1)} &concern, \textbf{importance}, \textbf{curiosity}, investment, \textbf{involvement}, attention, focussed, fascination, focus, \textbf{significance} \\
\multirow{1}{*}{BERT (L12)} &\textbf{curiosity}, \textbf{enthusiasm}, concern, fascination, confidence, unease, belief, \textbf{excitement}, desire, \textbf{passion}\\
\multirow{1}{*}{BERT (L24)} &appetite, attracting, attractiveness, actively, \textbf{demand}, attention, intrigued, \textbf{popularity}, \textbf{enthusiasm}, flocking\\
\midrule

\multirow{1}{*}{BART-Enc (L1)} &fascination, concern, intrigue, \textbf{involvement}, \textbf{relevance}, investment, stake, \textbf{curiosity}, \textbf{enthusiasm}, trustworthiness\\
\multirow{1}{*}{BART-Enc (L6)} &\textbf{enthusiasm}, fascination, appetite, \textbf{curiosity}, intrigued, \textbf{excitement}, intrigue, inclination, eagerness, enjoyment\\
\multirow{1}{*}{BART-Enc (L12)}&todo, aswell, inbetween, \textbf{enthusiasm}, fascination, eagerness, appetite, intrigued, attention, inclination\\\midrule
\multirow{1}{*}{BART-Dec (L1)} &fascination, \textbf{involvement}, intrigue, concern, participation, \textbf{curiosity}, investment, \textbf{excitement}, intrigued, \textbf{importance}\\

\multirow{1}{*}{BART-Dec (L6)} &fascination, appetite, \textbf{involvement}, engagement, affinity, uptake, \textbf{demand}, inclination, participation, appreciation\\

\multirow{1}{*}{BART-Dec (L12)} &participation, uptick, delight, decline, increase, spike, faith, decrease, grounding, surge\\

\bottomrule
\end{tabular}
\end{adjustbox}
\end{center}
\caption{Examples of  substitutes predicted by our method (w/o rerank) using different layers. Gold shows a list of ``conceivable'' words sorted by their annotated scores (with ``acceptable'' words shown in italic, and multiword expressions omitted from the table). Words included in Gold are boldfaced.}
\label{layer_examples}
\end{table*}

\section{Examples of Generated Substitutes}\label{layer_examples_section}
\Cref{layer_examples} shows examples of substitutes generated by our method using different layers (without reranking). It shows that the words retrieved by each layer are very different, indicating that each layer encodes very different information about the input word. For instance, given the target word \ex{care}, the first layer of BERT and BART-Enc/Dec retrieves a large number of words that contain the target word as a sub-morpheme (e.g.\ \ex{aftercare}, \ex{carefree}).\footnote{Since the edit distances between these words and the target word \ex{care} are not greater than the threshold (0.5), they weren't filtered out by our heuristic.} This is presumably because the first-layer representations are highly affected by the input word embedding, and hence result in retrieving words that share the same subword token (e.g.\ \ex{care \#\#free}) regardless of the semantic similarity. The last layer also performs poorly (as previously shown in \Cref{layer_fig}), e.g.\ BART-DEC (L12) retrieves \ex{participation} as the closest word to the target word \ex{interest}. This is because the last-layer representations of BART-decoder are used to directly predict the next word \ex{in} after \ex{interest} in the target sentence, and in fact, most of the retrieved words (e.g.\ \ex{uptick, faith, surge}) are those that often collocate with \ex{in}. Oddly, BART-Enc predicts a large number of substitutes that consist of multiple words (segmented by the tokeniser), none of which are relevant to the target word, e.g.\ \ex{aswell}, \ex{todo}, and \ex{inbetween} as substitutes for \ex{interest}. In fact, the number of such words increases (and the performance decreases) as the hyper-parameter $\lambda$ gets bigger (which increases the influence of $f(x,c)$ on the predictions). One possible interpretation is that the last layer representations of the BART encoder may contain vague contextual information rather than the lexical information of the input word, since they are used by the decoder to predict various words (esp.\ masked words) during pre-training. Lastly, another interesting observation is that, for the target word \ex{interest}, the last layer representations of BERT and BART-enc retrieve a lot of words that start from a vowel sound, despite the absence of the article \ex{an} before \ex{interest}, suggesting that the embeddings contain some morphophonetic information.
\end{document}